\documentclass{article}
\usepackage{arxiv}

\usepackage[utf8]{inputenc} % allow utf-8 input
\usepackage[T1]{fontenc}    % use 8-bit T1 fonts

\usepackage{graphicx}
\usepackage{amsmath}
\usepackage{hyperref}

\usepackage{booktabs} % For better-looking tables
\usepackage{siunitx} % For aligning numbers in tables
\usepackage{pgfplots}
\usepackage{tabularx} 

\usepackage{enumitem}
\usepackage{listings}
\usepackage{xcolor}
\usepackage{algorithm}
\usepackage{algpseudocode}
\usepackage{tikz}
\usepackage{subcaption}
\usepackage{adjustbox}
\usepackage{caption}
\usepackage{setspace}
\usepackage{makecell}
\usepackage{multirow}
\usepackage{url}

\definecolor{eclipseGreen}{RGB}{63, 127, 95}  
\algrenewcommand{\algorithmiccomment}[1]{\hfill \textcolor{eclipseGreen}{\footnotesize // #1}}

\AtBeginDocument{%
  }

\newcommand{\tool}{\textsc{Argus}}

\begin{document}

\title{Generating Realistic, Diverse, and Fault-Revealing Inputs with Latent Space Interpolation for Testing Deep Neural Networks}

\author{
 Bin Duan \\
  The University of Queensland \\ 
  %Brisbane, Australia \\
  \texttt{b.duan@uq.edu.au} \\
  \And
 Matthew B. Dwyer  \\
  University of Virginia\\
  \texttt{matthewbdwyer@virginia.edu}\\
 \And
 Guowei Yang \\
  The University of Queensland \\
  %Brisbane, Australia \\
  \texttt{guowei.yang@uq.edu.au} \\
}

\maketitle
\begin{abstract}

Deep Neural Networks (DNNs) have been widely employed across various domains, including safety-critical systems, necessitating comprehensive testing to ensure their reliability. Although numerous DNN model testing methods have been proposed to generate adversarial samples that are capable of revealing faults, existing methods typically perturb samples in the input space and then mutate these based on feedback from the DNN model. These methods often result in test samples that are not realistic and with low-probability reveal faults. To address these limitations, we propose a black-box DNN test input generation method, \tool, to generate realistic, diverse, and fault-revealing test inputs. \tool\ first compresses samples into a continuous latent space and then perturbs the original samples by interpolating these with samples of different classes. Subsequently, we employ a vector quantizer and decoder to reconstruct adversarial samples back into the input space. Additionally, we employ discriminators both in the latent space and in the input space to ensure the realism of the generated samples. 
Evaluation of \tool\, in comparison with state-of-the-art black-box testing and white-box testing methods, shows that \tool\ excels in generating realistic and diverse adversarial samples relative to the target dataset, and \tool\ successfully perturbs all original samples and achieves up to 4 times higher error rate than the best baseline method. Furthermore, using these adversarial samples for model retraining can improve model classification accuracy.

\end{abstract}

\section{Introduction}

Deep Neural Network~\cite{pinaya2020convolutional} (DNNs) have been widely applied in security-sensitive fields~\cite{ling2019deepsec} and safety-critical systems~\cite{forsberg2020challenges} such as autonomous driving~\cite{li2019stereo}, facial recognition~\cite{ben2021face}, and malware detection~\cite{ganesh2017cnn}. Like traditional software applications, these require extensive testing to ensure safe and reliable deployment~\cite{shahin2017continuous}. However, the mechanism of DNNs is significantly different from that of traditional software. Traditional software testing primarily focuses on identifying erroneous outcomes caused by code defects~\cite{wysopal2006art}. In contrast, during training, a DNN model learns to compute feature embeddings that allow it to classify data associated with different labels~\cite{chen2018understanding}. Due to the diversity of data classes, the features of different classes vary, making DNNs predict the class with the highest likelihood of matching the input features~\cite{leal2016learning}. When the input features are not distinct to those of other classes, DNNs may incorrectly predict another label~\cite{abdulnabi2015multi}. Such errors are not due to internal flaws in the DNN model, but arise because the training process may not include such error-prone inputs that are similar to other classes. 
In this way, the mechanism of DNN training limits the effectiveness of traditional software testing methods~\cite{thota2020survey}.

To ensure the predictive accuracy of DNNs, various testing methods~\cite{guo2018dlfuzz, odena2019tensorfuzz, lee2020effective, wang2022bet, pei2017deepxplore, dola2024cit4dnn, xie2019diffchaser, hu2023atom} have been proposed to identify model defects. 
When testing DNN models, the primary task is to generate samples that can reveal the erroneous behaviors of the model.
Current DNN model testing methods can be divided into white-box and black-box testing. In white-box testing, existing methods~\cite{lee2020effective, pei2017deepxplore, guo2018dlfuzz, xie2019deephunter} leverage neuron selection strategies to achieve high neuron coverage and identify samples that may lead to errors. %Although these methods can discover model flaws, 
They typically require internal knowledge of the target model, including its structure and parameters. Yet, the internal model information may not always be accessible in practice, especially in privacy-sensitive testing scenarios~\cite{keskin2021cyber, agahari2022not}.
In contrast, black-box testing methods~\cite{odena2019tensorfuzz, wang2022bet, hu2023atom, dola2024cit4dnn} do not require access to the internal details of the model. They generally identify model defects or inconsistencies by iteratively querying model outputs and updating test samples based on feedback. {However, many of these methods employ fuzzing~\cite{odena2019tensorfuzz, guo2018dlfuzz, wang2022bet} to generate adversarial samples, which tend to produce a large number of invalid samples that do not cause misclassifications, reducing the success rate of generating adversarial samples~\cite{dola2021distribution}. Moreover, even though perturbation has been shown to be effective in generating adversarial samples, most methods directly conduct perturbations in the input space~\cite{odena2019tensorfuzz, guo2018dlfuzz, pei2017deepxplore, lee2020effective, xie2019diffchaser, wang2022bet, hu2023atom} and thus may generate samples whose features do not correspond to those represented by the original class.}

%\bin{However, in privacy-sensitive testing scenarios~\cite{keskin2021cyber, agahari2022not}, white-box testing cannot access internal model information. In contrast, black-box testing methods~\cite{febriyanti2021implementasi} do not require access to the internal details of the model, making them more suitable for secure testing scenarios. Black-box DNN model testing methods ~\cite{odena2019tensorfuzz, wang2022bet, xie2019diffchaser} generally identify model defects or inconsistencies by iteratively querying model outputs and updating test samples based on feedback. {However, (a) existing DNN testing methods employ fuzzing techniques~\cite{zhu2022fuzzing} to generate adversarial samples, which tend to produce a large number of invalid samples that do not cause misclassifications, reducing the success rate of generating effective adversarial samples~\cite{dola2021distribution}. Additionally, (b) these methods that directly conduct perturbations in the input space may generate samples whose features do not correspond to those represented by the original class, further exacerbating the limitations of sample generation in black-box testing.} }

We contend that for black-box DNN test input generation methods to be usable in practice, they must
generate samples that are: \textbf{realistic} and \textbf{diverse} relative to a target dataset, i.e.,
the dataset on which the model under test was trained;
and that with high-probability \textbf{reveal faults} in models trained to state-of-the-art accuracy.
Moreover, whereas white-box test generation approaches must generate new tests for each version of the
model under test across its development cycle, for black-box testing the cost of test generation 
is less of a concern since new tests need only be generated when the underlying dataset shifts.
These properties of black-box test generation approaches help to 
(a) ensure a low false positive rate -- since realistic fault-revealing inputs are one's developers will be concerned about; 
(b) better utilize testing efforts -- since diverse fault-revealing
inputs allow testers to minimize the triage of redundant tests and ensure a test run covers a breadth of behavior; and
(c) increase the probability of a test sample causing an error -- since a high-probability of failure means
fewer tests need to be generated in order to reveal model faults.

In this work, we propose \tool, that aims to generate realistic, diverse, and fault-revealing inputs to test DNN models. {These adversarial samples are not only closely similar to the original samples but also effectively mislead the model's predictions, causing classification errors across multiple classes.}. Specifically, \tool\ utilizes, adapts, and extends a VQ-VAE~\cite{van2017neural}, a generative model where an encoder compresses complex input samples, images, into a low-dimensional latent space, allowing the model to learn the features of the training data and obtain a suitable representation for reconstruction. After obtaining the latent representation of the sample, we introduce perturbations by interpolating with the latent representations of samples from other classes, using a perturbation factor $\lambda$ to control the degree of the perturbation. The perturbed latent representation is then
reconstructed back to a sample in the input space.

Direct perturbations in the input space often lead to reduced realism and a large number of samples that fail to reveal model faults~\cite{berend2020cats,dola2021distribution,riccio2023and}. 
Unlike previous DNN model testing methods that directly perturb images in the input space, \tool\ perturbs in the latent space, ensuring that the generated samples are realistic. Since these perturbations occur in the latent space, they are highly inconspicuous, and the generated adversarial samples exhibit slight differences from the original samples.
Moreover, \tool\ %overcomes these deficiencies by 
incorporates discriminators, which are trained to force the perturbation process to remain close to the data distribution, which significantly improves the realism of generated samples. Additionally, to enhance the perturbation success rate and induce diverse classification errors, we introduce features from different classes by interpolating the latent representations of other classes. This latent space interpolation perturbation method leverages the fact that classification models output the class with the highest probability. By introducing {a small portion of} features from other classes in the latent space, the generated samples contain features of both the original and other classes, thereby increasing the likelihood of misclassification. Consequently, the adversarial samples generated by \tool\ not only retain realism relative to the original samples but also exhibit a higher error rate and expose a diversity of failure modes.

We conducted an extensive evaluation of \tool, on three widely used datasets, MNIST~\cite{lecun2010mnist}, CIFAR10~\cite{krizhevsky2009learning}, and ImageNet~\cite{deng2009imagenet} using six well-known DNN models: LeNet4, LeNet5~\cite{lecun1998gradient}, VGG16, VGG19~\cite{simonyan2014very}, ResNet18, and ResNet50~\cite{he2016deep}. 
We compared \tool, with state-of-the-art black-box testing and white-box testing methods. 
The results show that, with the same number of generated samples, the adversarial samples generated by \tool\ outperform baseline methods in several aspects: the rate at which they elicit errors in model behavior is up to 4 times higher than the best baseline method, the realism of the generated samples is significantly better than all baseline methods, and the perturbation success rate reaches 100\%. Additionally, in terms of sample diversity, \tool\ can exceed all baseline methods. Furthermore, using these adversarial samples for model retraining can improve model classification accuracy. 
%by approximately 1\%-3\%.

In summary, our contributions are as follows.
\begin{itemize}
    \item \textbf{Approach}.  We propose a novel black-box testing method for DNNs named \tool, which effectively generates high-realism adversarial samples through interpolating perturbations in a sample's latent representation. This method can mislead model classifications and cover diverse output labels.
    \item \textbf{Tool}. We implement \tool\ in PyTorch, including the components for sample generation through the combination of VQ-VAE and dual discriminators, and have made the code publicly available upon acceptance.
    \item \textbf{Evaluation}. We evaluate \tool\ in comparison to the state of the art, demonstrating that \tool\ outperforms existing black-box and white-box testing methods in various aspects. Our method achieves an error rate of up to approximately 62\%, 100\% adversarial success rate, generates adversarial samples that are more realistic relative to the original dataset, and triggers errors in diverse class labels.
\end{itemize}

\begin{figure}[t!]
  \centering
  \includegraphics[width=1\linewidth]{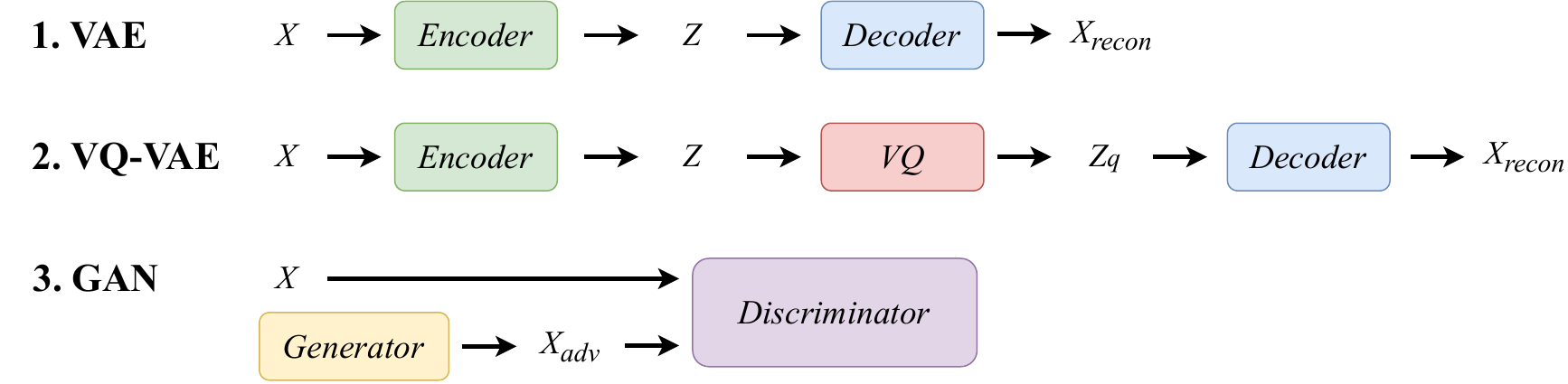}
  \caption{VAE, VQ-VAE, and GAN Architecture}
  \label{fig:vae-vq-gan}
\vspace{-5mm}
\end{figure}

\section{Background}

\subsection{VQ-VAE}
\label{sec:pretrain}

Extensive research in the field of machine learning has demonstrated that manipulating the latent space~\cite{bengio2013representation, gat2022latent, shen2020interpreting}, a low-dimensional embedding specifically designed to capture key variational factors within observed data, is not only more efficient but also more covert compared to direct manipulations in the input space~\cite{burgess2018understanding}. This capability to manipulate the latent space uniquely positions generative models to effectively utilize these latent representations to generate samples that are similar to, but distinct from, the training data.

Variational Autoencoders (VAE)~\cite{kingma2013auto} are classic generative models comprising an \(Encoder\) and \(Decoder\), as shown in Fig~\ref{fig:vae-vq-gan}. The \(Encoder\) maps input samples \(X\) to continuous latent vectors \(Z\), which the \(Decoder\) reconstructs into \(X_{{recon}}\). Modern VAEs achieve superior reconstruction across domains. The Vector Quantized Variational Autoencoder (VQ-VAE)~\cite{van2017neural}, shown in Fig~\ref{fig:vae-vq-gan}, enhances VAE by introducing a \(Vector\ Quantizer\)~(VQ), converting \(Z\) into discrete representations \(Z_{{q}}\), improving data representation and reconstruction~\cite{yan2021videogpt}. This method enhances performance on complex data while maintaining efficiency. The training process is as follows:

The input sample \(X\) is entered into \(Encoder\) to obtain the continuous latent representation \(Z\):

\begin{equation}
Z = \text{\(Encoder\)}(X)
\label{eq:encoder}
\end{equation}

This enables the \(Encoder\) to learn key latent representations, extracting essential features that provide comprehensive information for image reconstruction.

\( Z\) is quantized using a learnable codebook to produce a discrete latent representation \( Z_{q} \):

\begin{equation}
Z_{{q}} = \arg\min_{c} \| Z - c \|^2, \quad c \in \text{Codebook}
\label{eq:quantization}
\end{equation}

The quantization step discretizes the latent space into a learnable codebook, enabling efficient and high-quality reconstruction. Specifically, \(Vector\ Quantizer\) maps each latent vector \(Z\) to the nearest code, yielding a discrete representation \(Z_q\). This compression reduces complexity, ensures consistent inputs for the \(Decoder\), and prevents posterior collapse in VAEs, preserving meaningful latent representations. 
By bridging continuous and discrete representations, vector quantization enhances the model’s ability to learn robust and meaningful features for high-quality generation.

The quantized latent representation \( Z_{{q}} \) is then entered into \(Decoder\) to reconstruct the input sample \( X_{{recon}} \):

\begin{equation}
X_{{recon}} = \text{\(Decoder\)}(Z_{{q}})
\label{eq:decoder}
\end{equation}

Here, the optimization target is the entire generative model, including \(Encoder\), \(Vector\ Quantizer\), and \(Decoder\), using a loss function that ensures high-quality sample reconstruction. The loss function is defined as:

\begin{equation}
\mathcal{L} = \mathcal{L}_{{rec}} + \alpha \mathcal{L}_{{cb}} + \beta \mathcal{L}_{{com}}
\label{eq:total_loss}
\end{equation}

Where, \( \mathcal{L}_{{rec}} \) is the reconstruction loss, which measures the difference between the input sample \( X \) and the reconstructed sample \( X_{{recon}} \):

\begin{equation}
\mathcal{L}_{{rec}} = \| X - X_{{recon}} \|^2
\label{eq:reconstruction_loss}
\end{equation}

\( \mathcal{L}_{{cb}} \) is codebook loss, which encourages \(Encoder\)'s outputs to be close to the nearest codebook entries, enhancing the effectiveness of the codebook:

\begin{equation}
\mathcal{L}_{{cb}} = \| Z - \text{sg}(Z_{{q}}) \|^2
\label{eq:codebook_loss}
\end{equation}

\( \mathcal{L}_{{com}} \) is commitment loss, which ensures the stability of the quantized representations by penalizing deviations:

\begin{equation}
\mathcal{L}_{{com}} = \| \text{sg}(Z) - Z_{{q}} \|^2
\label{eq:commitment_loss}
\end{equation}

In these equations, \( \text{sg}(\cdot) \) denotes the stop-gradient operator, preventing gradient flow through its argument during backpropagation. The hyperparameters \( \alpha \) and \( \beta \) balance the codebook and commitment losses. VQ-VAE training aims to minimize the reconstruction error between \( X \) and \( X_{{recon}} \) while aligning the latent space distribution with a standard normal distribution.

\subsection{GAN}
A {Generative Adversarial Network (GAN)~\cite{goodfellow2020generative}, as shown in Fig.~\ref{fig:vae-vq-gan}, consists of two components: a \(Generator\) and a \(Discriminator\), trained adversarially. The \(Generator\) produces adversarial samples \(X_{{adv}}\) mimic real samples \(X\), while the \(Discriminator\) learns to differentiate between real and adversarial samples. As the training progresses, the \(Generator\) improves its ability to deceive the \(Discriminator\), while the \(Discriminator\) becomes better at identifying fake samples. This competition drives the \(Generator\) to produce increasingly realistic adversarial samples. 

The training process uses a composite loss function that balances two objectives: the \(Discriminator\)'s aims to correctly classify real samples \(X\) as real while identifying generated adversarial samples \(X_{{adv}}\) as fake, while the \(Generator\) strives to generate adversarial samples that can fool the \(Discriminator\). The \(Discriminator\)'s loss is defined as follows:  

\begin{equation}
\mathcal{L}_{D} = -\left[ \log {D}(X) + \log \left( 1 - {D}(X_{{adv}}) \right) \right]
\label{eq:lossD}
\end{equation}

The \(Generator\)'s loss \( \mathcal{L}_{G} \) is designed to maximize the likelihood that the \(Discriminator\) classifies \(X_{{adv}}\) as real:

\begin{equation}
\mathcal{L}_{G} = -\log {D}(X_{{adv}})
\label{eq:lossZG}
\end{equation}

Here, \( D(\cdot) \) represents the \(Discriminator\)'s output. }

This paper introduces a dual-GAN architecture that operates in both latent space and input spaces to enhance the realism of adversarial samples while incorporating perturbations. 

\section{Approach}

\begin{figure*}[t!]
  \centering
  \includegraphics[width=0.8\linewidth]{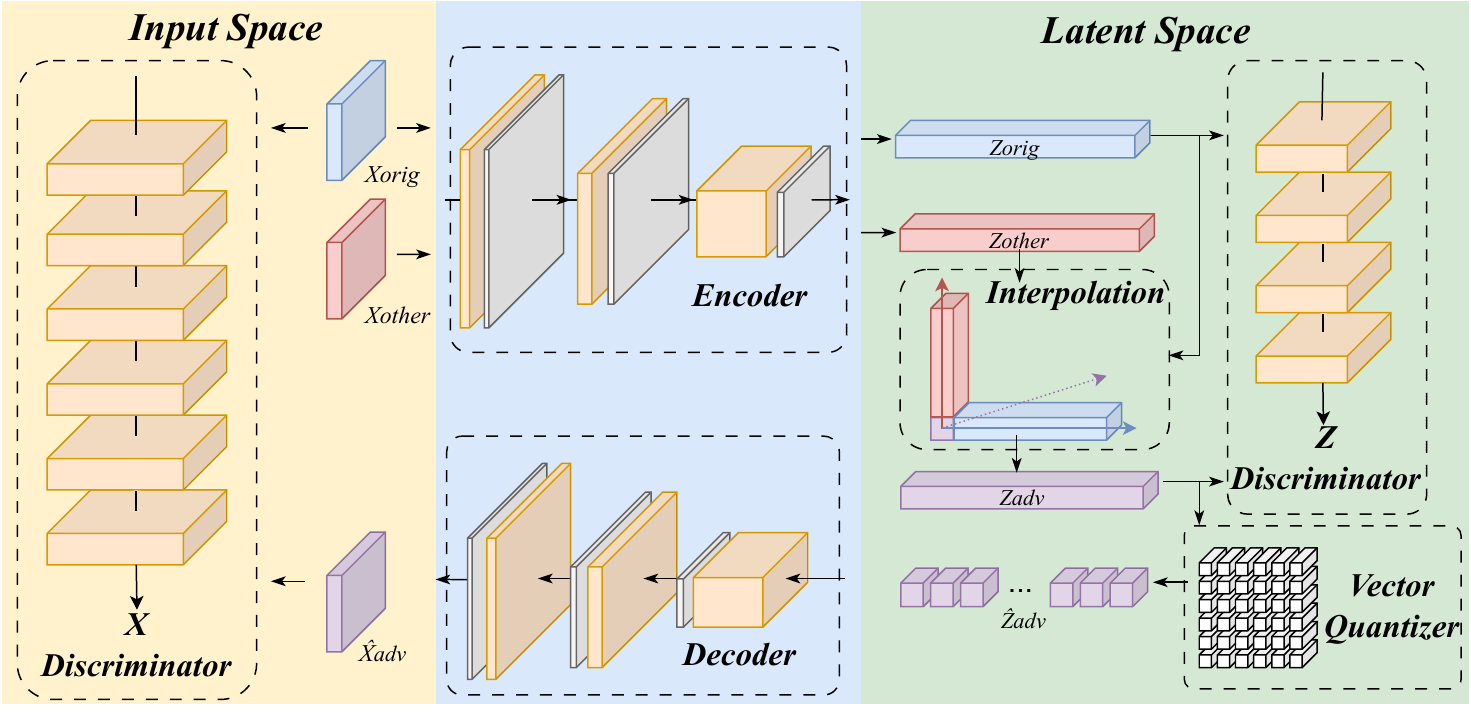}
  \vspace{-4mm}
  \caption{{Overall Framework of \tool}}
  \label{fig:overview}
\vspace{-4mm}
\end{figure*}

{
In this section, we present \tool, a novel black-box testing method for DNN models. \tool\ generates adversarial samples by interpolating the latent representation of an original sample with those of samples from other classes.  Fig~\ref{fig:overview} shows the overall framework of \tool. The framework begins by training a VQ-VAE to achieve high-quality image reconstruction. \tool\ then uses the VQ-VAE encoder to map both the original and other-class samples into a low-dimensional latent space, where adversarial latent representations are created through interpolation.  
To enhance the realism of adversarial samples, \tool\ employs a two-stage generation process. First, a latent-space discriminator refines the adversarial latent representation, reducing perturbation-induced differences. Next, the VQ-VAE’s vector quantizer converts the adversarial latent representation into a discrete form, which is then decoded back into the input space to reconstruct the adversarial sample. Finally, an input-space discriminator further aligns the adversarial sample with the original sample to enhance its realism.

}

\subsection{Training VQ-VAE with Discriminators}
\label{sec:3.1}

{We first pre-train VQ-VAE to ensure effective information compression and recovery, which is essential for generating high-quality samples. As detailed in Sec~\ref{sec:pretrain}, once the VQ-VAE pre-training is complete, we proceed with  joint training with discriminators to enhance its ability to generate more realistic adversarial samples that closely resemble the original samples.}

\subsubsection{Generating adversarial samples in the latent space}
\label{sec:3.1.1}

{We generate adversarial latent representations by applying interpolation-based perturbations in the continuous latent space. As shown in Fig.~\ref{fig:overview}, we utilize the pre-trained VQ-VAE modules: \(Encoder\), \(Vector\ Quantizer\) and \(Decoder\). The original sample \(X_{{orig}}\) and the sample from another class \(X_{{other}}\) are fed into the \(Encoder\), which converts them to continuous latent representations \(Z_{{orig}}\) and \(Z_{{other}}\), respectively.}

\begin{equation}
Z_{{orig}} = \text{\(Encoder\)}(X_{{orig}}), Z_{{other}} = \text{\(Encoder\)}(X_{{other}})
\label{eq:encoder_other}
\end{equation}

Then, by interpolating between \(Z_{{orig}}\) and \(Z_{{other}}\), the adversarial latent representation \(Z_{{adv}}\) is obtained. Specifically, interpolation is performed by taking a weighted combination of \(Z_{{orig}}\) and \(Z_{{other}}\), where a factor \(\lambda \in [0, 1]\) should be the contribution of each latent representation to the final result. The interpolation can be expressed as:

\begin{equation}
Z_{{adv}} = \lambda Z_{{other}} + (1 - \lambda) Z_{{orig}}, \quad \lambda \in [0, 1]
\label{eq:interpolation}
\end{equation}

{Here, \( \lambda \) acts as a perturbation factor, controlling the influence of the other class's latent representation. By adjusting \(\lambda\), we can regulate the degree of perturbation applied to the original latent representation. This process incorporates a small portion of features from different classes into the adversarial sample, increasing the likelihood of model misclassification while preserving realism relative to the original sample. 
Unlike direct input space perturbations, interpolation in the latent space ensures that the resulting adversarial sample retains natural sample properties, as the latent space captures abstract and meaningful features.}

After obtaining \(Z_{{adv}}\), we introduce \(Z\ Discriminator\) in the latent space to ensure the realism of the generated adversarial samples. {At this stage, \(Encoder\) acts as the generator for \(Z_{{orig}}\), while \(Encoder\) combined with the interpolation operation acts as the generator for \(Z_{{adv}}\). \(Z\ Discriminator\) receives both \(Z_{{orig}}\) and \(Z_{{adv}}\) as inputs, enabling it to distinguish between the original latent representation and the adversarially interpolated latent representation.}

In this adversarial training phase in the latent space, we aim to optimize \(Encoder\) and \(Z\ Discriminator\). We employ a composite loss function that balances \(Z\ Discriminator\)'s ability to correctly classify \( Z_{{orig}} \) as belonging to the original sample and correctly identify \( Z_{{adv}} \) as adversarial samples, alongside the generator's objective to generate adversarial representations that can fool \(Z\ Discriminator\). {The discriminator loss is defined as:}

\begin{equation}
\mathcal{L}_{D_z} = -\left[ \log {D_z}(Z_{{orig}}) + \log \left( 1 - {D_z}(Z_{{adv}}) \right) \right]
\label{eq:lossZD}
\end{equation}

The generator loss \( \mathcal{L}_{G_z} \) is designed to maximize \(Z\ Discriminator\)'s probability of misclassifying \( Z_{{adv}} \) as originating from the original sample distribution:

\begin{equation}
\mathcal{L}_{G_z} = -\log {D_z}(Z_{{adv}})
\label{eq:lossZG}
\end{equation}

\({D_z}(\cdot)\) is \(Z\ Discriminator\)'s output, which estimates the probability that the given latent representation is from the original sample. {The loss function \( \mathcal{L}_{D_z} \) consists of two terms: (1) \(-\log {D_z}(Z_{{orig}})\) encourages \(Z\ Discriminator\) to correctly classify \(Z_{{orig}}\) as belonging to the original sample; and (2) \(-\log (1 - {D_z}(Z_{{adv}}))\) encourages \(Z\ Discriminator\) to correctly identify \(Z_{{adv}}\) as adversarial sample. 
On the other hand, the loss function \( \mathcal{L}_{G_z} \) is designed to minimize the difference between \(Z_{{adv}}\) and \(Z_{{orig}}\), ensuring that the adversarial latent representation closely resemble the original latent representation. This adversarial training process enhances the quality of \(Z_{{adv}}\), making it increasingly realistic and harder for the discriminator to distinguish from \(Z_{{orig}}\).}

{In this process, the \(Encoder\)  and interpolation generate \(Z_{{adv}}\), which is designed to be challenging for \(Z\ Discriminator\) to distinguish from \(Z_{{orig}}\), while \(Z\ Discriminator\)  continuously improves its ability to differentiate between \(Z_{{adv}}\) and \(Z_{{orig}}\). This adversarial mechanism forces \(Z_{{adv}}\) to increasingly resemble \(Z_{{orig}}\), which is critical because interpolation may cause latent representations to deviate from the original data distribution, potentially leading to unrealistic samples. This is relevant during interpolations between different classes, which might generate ambiguous or indistinct representations. \(Z\ Discriminator\) plays a key role in constraining these interpolated latent representations, ensuring that they remain realistic.}

\subsubsection{Reconstructing adversarial samples in the input space}
\label{sec:3.1.2}
As shown in Fig.~\ref{fig:overview}, after obtaining \(Z_{{adv}}\),
%which approximates the original sample
the \(Vector\ Quantizer\) is used to convert the continuous adversarial sample \(Z_{adv}\) into a discrete adversarial sample \(\hat{Z}_{adv}\). 

\begin{equation}
\hat{Z}_{adv}= \arg\min_{c} \| Z_{{adv}} - c \|^2, \quad c \in \text{Codebook}
\label{eq:quantization_adv}
\end{equation}

{Subsequently, \(Decoder\) is used to reconstruct \(\hat{Z}_{adv}\) back into the input space, generating the reconstructed adversarial sample \(\hat{X}_{{adv}}\). }

\begin{equation}
\hat{X}_{{adv}} = \text{\(Decoder\)}(\hat{Z}_{adv})
\label{eq:decoder}
\end{equation}

\(X_{{orig}}\) and \(\hat{X}_{{adv}}\) are fed into \(X\ Discriminator\), which is trained to distinguish between the original and adversarial samples. Similar to \(Z\ Discriminator\), the discriminator's loss function aims to differentiate between the original and adversarial samples.
\(X\ Discriminator\) loss is given by:
\begin{equation}
\mathcal{L}_{D_x} = -\left[ \log {D_x}(X_{{orig}}) + \log \left( 1 - {D_x}(\hat{X}_{{adv}}) \right) \right]
\label{eq:lossXD}
\end{equation}

{The generator loss \( \mathcal{L}_{G_x} \) is designed to maximize \(X\ Discriminator\)'s probability of misclassifying the asversarial sample \(\hat{X}_{{adv}}\) as the original sample \(X_{{orig}} \). This encourages the generator to produce adversarial samples that are indistinguishable from the original samples. The loss function can be expressed as:}
\begin{equation}
\mathcal{L}_{G_x} = -\log {D_x}(\hat{X}_{{adv}})
\label{eq:lossXG}
\end{equation}

{where, \( {D_x}(\cdot) \) represents \( X\ Discriminator\)'s output, estimating the probability that the input image is classified as real.  The discriminator's loss function \( \mathcal{L}_{D_x} \) consists of two terms: (1) \( -\log {D_x}(X_{orig}) \) encourages \( X\ Discriminator\) to correctly classify \( X_{orig} \) as a real (original) sample; and (2) \( -\log \) \(\left( 1 - {D_x}(\hat{X}_{{adv}}) \right) \) encourages the \( X\ Discriminator\) to correctly identify \(\hat{X}_{{adv}}\) as an adversarial sample. On the other hand, the generator loss \( \mathcal{L}_{G_x} \) aims to make \(\hat{X}_{{adv}}\) as close as possible to \( X_{orig} \), making it difficult for \( X\ Discriminator\) to distinguish between them.

The functions of \( X \ Discriminator\) are threefold:
%, is to distinguish between \( X \) and \(\hat{X}_{{adv}}\).
(1) it forces \(Encoder\), \(Vector\ Quantizer\), and \(Decoder\), to generate adversarial samples that are more realistic and indistinguishable from the original samples; (2) it helps maintain consistency in the data distribution, preventing the model from learning invalid  features; (3) working in tandem with \( Z \ Discriminator\), it enables simultaneous adversarial training in both latent and input spaces, achieving collaborative optimization. 

Additionally, during the training process, to preserve the reconstruction capability of the VQ-VAE, we incorporate the loss function from Equation~\ref{eq:total_loss} into the optimization process. This enables \(Encoder\) to effectively cluster data from different classes during encoding while retaining its robust feature extraction capability. This supports the subsequent reconstruction process performed by \(Vector\ Quantizer\) and \(Decoder\), maintaining the overall quality and realism of the generated samples.}

\subsection{Generating Adversarial Samples}
\label{sec:3.2}

\begin{algorithm}[t]
\setstretch{1.1}
\footnotesize
\caption{Generating Adversarial Samples with \tool}
\label{alg:generate_adversarial}
\begin{algorithmic}[1]
\State \textbf{Input:} Dataset $\mathcal{D}$ with $K$ classes, perturbation factor $\lambda$
\State \textbf{Output:} Adversarial sample images $\mathcal{D}_{adv}$
\State $\mathcal{D}_k = \{ X \mid \text{label}(X) = k, k \in [0, K-1]$ \} 
\Comment{$\mathcal{D}$ split into $K$ subsets $\mathcal{D}_k$. }
\State $\mathcal{X}_{orig}^k= \text{Select}\left(\mathcal{D}_k\right)$ 
\Comment{Select original samples from each class $k$.}
\State $\mathcal{X}_{other}^k = \text{Select}\left(\mathcal{D}_k\right)$ 
\Comment{Select perturbed samples from each class $k$.}
%\State $\mathcal{X}_{orig} = \bigcup_{k=0}^{K-1} \mathcal{X}_{orig}^k $
%\State $\mathcal{X}_{other} = \bigcup_{k=0}^{K-1} \mathcal{X}_{other}^k $
\State $\mathcal{Z}_{{orig}}^k = \text{\(Encoder\)}(\mathcal{X}_{{orig}}^k)$
\State $\mathcal{Z}_{{other}}^k = \text{\(Encoder\)}(\mathcal{X}_{{other}}^k)$
\State Initialized empty set $\mathcal{D}_{adv}$
\For{each class $k$}
  \For{$Z_{orig}^{(k, i)}$ in $\mathcal{Z}_{orig}^k$} \Comment{($k, i$), $i$-th sample in $k$-th class.}
    \For{each class $k'$}
      \For{$Z_{other}^{(k', j)}$ in $\mathcal{Z}_{other}^k$  ($k' \neq k$)} \Comment{($k', j$), $j$-th sample in $k'$-th class.}
                \State $Z_{adv}^{(k, i, j)} = \lambda Z_{other}^{(k', j)} + (1 - \lambda) Z_{orig}^{(k, i)}$
                \State $\hat{Z}_{adv}^{(k, i, j)} = \text{\(VectorQuantizer\)}(Z_{adv}^{(k, i, j)})$
                \State $\hat{X}_{adv}^{(k, i, j)} = \text{\(Decoder\)}(\hat{Z}_{adv}^{(k, i, j)})$ 
                \State $\mathcal{D}_{adv} = \mathcal{D}_{adv} \cup \{\hat{X}_{adv}^{(k, i, j)} \}$
      \EndFor
    \EndFor
  \EndFor
\EndFor
\State \textbf{Return} $\mathcal{D}_{adv}$
\end{algorithmic}
\end{algorithm}

%\vspace{-4mm}

{After completing the joint training of VQ-VAE and discriminators, the model is ready to generate adversarial samples. Thanks to the adversarial training process in both the latent and input spaces, \(Encoder\), \(Vector\ Quantizer\), and \(Decoder\) have been optimized to generate samples that closely resemble the original samples. Therefore, the two discriminators are no longer needed for the adversarial sample generation phase. }

{Algorithm~\ref{alg:generate_adversarial} shows the main steps to generate adversarial samples by interpolating between the original and different classes of samples in the latent space.} This generation process begins with a given dataset, denoted as $\mathcal{D}$, and a specified perturbation factor $\lambda$~(line 1). The dataset $\mathcal{D}$ is first divided into different classes by labels, and a subset of both original and cross-class samples is selected~(lines 3-5). Following this, these samples are encoded into the latent space~(lines 6-7), where interpolation is performed to create adversarial representations~(lines 9-13). {It is worth noting that the $\lambda$ value used during generation phase should be consistent with the $\lambda$ value used during training, which ensures that the generated adversarial samples maintain the same level of realism and effectiveness as those generated during the training process. After the interpolation in the latent space is performed, the interpolated latent representation is quantized and decoded back into an adversarial sample~(lines 13-15),} forming $\mathcal{D}_{adv}$~(line 16), the final collection of adversarial samples for testing and improving model performance.

\section{{Evaluation}}
\subsection{Research Questions}
Our aim is to investigate the following research questions:
\begin{itemize}
  \item[\textbf{RQ1:}] How realistic are \tool\ generated samples?
  \item[\textbf{RQ2:}] How diverse are \tool\ generated samples?
  \item[\textbf{RQ3:}] How effective are \tool\ generated samples in improving model performance?
  \item[\textbf{RQ4:}] How effective are perturbations in \tool\ in modifying model output?
  \item[\textbf{RQ5:}] Does each of the main components of \tool\ contribute to its effectiveness?
\end{itemize}

To evaluate the effectiveness of \tool, we designed internal and external comparison experiments. The internal comparison experiment investigated the impact of the perturbation factor $\lambda$ on the outcomes of the samples to understand the significance and role of this factor in the sample generation process. The external comparison experiment is used to compare with the baseline methods.
For RQ1, we evaluate the similarity between the generated samples and the original samples. Our goal is to make the generated samples more realistic and closer to the original ones, rather than exhibiting significant deviations. To achieve this, we design experimental metrics to measure the differences between the original and generated samples. Additionally, for a more in-depth comparison, we conduct a visual evaluation.
For RQ2, we explore the diversity of the samples generated by \tool. Specifically, we apply samples from different label perturbations to the same original images in the latent space to observe if the generated samples can induce various classes, including perturbing the original images to cause different classifications. 
For RQ3, we explore the effectiveness of the generated samples by comparing the test accuracy of the DNN models before and after retraining with the generated samples. 
For RQ4, we evaluated it from two aspects: the success rate of perturbations and the proportion of generated erroneous images. 
For RQ5, to validate the contributions of each component within \tool, we performed ablation experiments. We compared the use of a VAE generative model, instead of VQ-VAE, and scenarios where no discriminator was employed, aiming to explore the necessity of each critical component.

\subsection{Experimental Setup}
\label{experimental}
\noindent\textbf{Target DNN Models and Datasets.} 
Following ~\cite{xie2019diffchaser, wang2022bet, lee2020effective, guo2018dlfuzz}, we selected three of the most widely used datasets and six of the most widely adopted DNN models for our experiments.

MNIST~\cite{lecun2010mnist} contains $28\times28$ grayscale images of handwritten digits; it has 60k training inputs and 10k test inputs. We train LeNet-4 and LeNet-5~\cite{lecun1998gradient} on this dataset.

CIFAR10~\cite{krizhevsky2009learning} contains $32\times32$ color images of objects belonging to 10 classes; it has 50k training inputs and 10k test inputs. We used pre-trained VGG16~\cite{simonyan2014very} and ResNet18~\cite{he2016deep} and fine-tuned this dataset. 

ImageNet~\cite{deng2009imagenet} contains $224\times224$ color images across 1000 classes, with 1.2 million training images and 50k validation images. We used pre-trained VGG19~\cite{simonyan2014very} and ResNet50~\cite{he2016deep} models on this dataset as our target DNNs. To facilitate comparison and account for computational resources, we followed~\cite{wang2022bet} and selected the first 50 classes of ImageNet (based on their numerical order) for our experiments.

\noindent\textbf{Framework.} \tool\ is implemented using PyTorch (v2.2.2) along with essential libraries for data processing, training, and evaluation. 

\noindent\textbf{Environment.} Our experiments were conducted on a 64-core PC with 32GB RAM and an NVIDIA RTX A6000 GPU. All of the test suites generated across all of our datasets were produced within three hours.

\noindent\textbf{Hyperparameters.} Following~\cite{van2017neural}, for the VQ-VAE generative model, training was conducted for 50 epochs, with an initial learning rate of 0.001. The learning rate decayed every 20 epochs, with a decay factor of 0.1. In the Formula~\ref{eq:total_loss}, \(\alpha = 0.25\) and \(\beta = 0.75\). For the latent space discriminator, the input dimension matches the output dimension of the VQ-VAE encoder, followed by three fully connected layers, reducing the output dimension to 1. For the input space discriminator, the input dimension corresponds to the original image size, processed through five convolutional layers and two fully connected layers, resulting in an output dimension of 1.

\noindent\textbf{Baselines.} We compare \tool\ with four open-source state-of-the-art testing methods DiffChaser~\cite{xie2019diffchaser}, ADAPT~\cite{lee2020effective}, DLFuzz~\cite{guo2018dlfuzz} and DeepXplore~\cite{pei2017deepxplore}. We randomly selected 10 original image samples in each data set and selected 30,000 perturbed samples for each adversarial attack of the original image. For MNIST and CIFAR10, each dataset contains 10 classes, so we selected approximately 3,333~(30,000/9) images from each class (excluding the class of the original sample). For ImageNet, since we selected the 50 priority classes, we chose approximately 612~(30,000/49) images from each class as adversarial samples.
For DiffChaser, we obtained the open-source code. Specifically, DiffChaser generates samples that lead to conflicting predictions between two versions of a model. Therefore, in the process of generating samples, we used the original model and its quantized version as the two versions for comparison. We employed 8-bit quantized models, as their paper shows that 8-bit quantized models exhibit better performance. For instance, we generated adversarial samples between LeNet5 and its 8-bit quantized version, LeNet5-8bit. Finally, these adversarial samples were tested on the original model. 
For Adapt, DLFuzz, and DeepXplore, we acquired their source codes from the Adapt open-source repository and successfully ran the experiments.
In the comparison experiments with the above baseline, we performed 30,000 perturbs for each original sample, and calculated the average results based on these generated adversarial samples.
BET~\cite{wang2022bet} is a black-box testing method for DNNs but it is closed source, so we are not able to compare \tool\ with it empirically.  We leave the empirical comparison to BET for future work, if that tool is made available to the community.

\subsection{Metrics}
\label{metrics}
\noindent\textbf{Mean Squared Error (MSE)}: Measures the average squared difference in pixel intensities between two images, quantifying similarity. A lower MSE indicates higher similarity~\cite{sara2019image}.  

\noindent\textbf{Structural Similarity Index Measure (SSIM)}: Assesses image similarity based on visual perception, considering luminance, contrast, and structure. SSIM ranges from -1 to 1, with 1 indicating perfect similarity. Unlike MSE, SSIM captures structural differences~\cite{sara2019image}.  

\noindent\textbf{Label Diversity (LD)}: Counts the number of distinct labels for which adversarial samples are found, reflecting the ability to explore different decision boundaries~\cite{wang2022bet, xie2019diffchaser, lee2020effective}.  

\noindent\textbf{Improved Classification Accuracy (ICA)}: Evaluates whether generated samples help cover previously misclassified images and improve classification accuracy through fine-tuning, following BET~\cite{wang2022bet}.  

\noindent\textbf{Error Rate (ER)}: Represents the proportion of perturbations that cause misclassification, indicating the perturbation method’s effectiveness in exposing model weaknesses.  

\noindent\textbf{Success Rate (SR)}: Measures the percentage of original samples successfully generating adversarial counterparts. A higher SR indicates a more effective testing method~\cite{wang2022bet, xie2019diffchaser, lee2020effective}.  

\begin{table}[h]
\centering
\caption{Results of Comparing Samples Generated by Different Methods.}
\tiny
\label{sim}
\setlength{\tabcolsep}{0.5pt} % 将列间距设置为3pt
\begin{tabular}{@{}lccc|ccc|ccc|ccc|ccc|ccc@{}}
\toprule
& \multicolumn{6}{c}{MNIST} & \multicolumn{6}{c}{CIFAR10} & \multicolumn{6}{c}{ImageNet}\\
\cmidrule(lr){2-7} \cmidrule(lr){8-13} \cmidrule(lr){14-19}
& \multicolumn{3}{c}{LeNet4} & \multicolumn{3}{c}{LeNet5} & \multicolumn{3}{c}{VGG16} & \multicolumn{3}{c}{ResNet18} & \multicolumn{3}{c}{VGG19} & \multicolumn{3}{c}{ResNet50} \\
\cmidrule(lr){2-4} \cmidrule(lr){5-7} \cmidrule(lr){8-10} \cmidrule(lr){11-13} \cmidrule(lr){14-16} \cmidrule(lr){17-19}
Method    & MSE    & SSIM   & LD  & MSE    & SSIM   & LD   & MSE    & SSIM   & LD  & MSE    & SSIM   & LD  & MSE    & SSIM   & LD  & MSE    & SSIM   & LD  \\
\midrule
\tool-0.1   & \colorbox{gray!50}{\textbf{0.0011}} & \colorbox{gray!50}{\textbf{0.9869}} & {3.1} & \colorbox{gray!50}{\textbf{0.0012}} & \colorbox{gray!50}{\textbf{0.9902}} & \colorbox{gray!50}{3.6} & \colorbox{gray!50}{\textbf{0.0035}} & \colorbox{gray!50}{\textbf{0.9247}} & {4.1} & \colorbox{gray!50}{\textbf{0.0036}} & \colorbox{gray!50}{\textbf{0.9152}} & {4.4} & \colorbox{gray!50}{\textbf{0.0050}} & \colorbox{gray!50}{\textbf{0.8801}} & {16.0} & \colorbox{gray!50}{\textbf{0.0051}} & \colorbox{gray!50}{\textbf{0.8992}} & {18.2} \\
\tool-0.2   & \colorbox{gray!50}{{0.0015}} & \colorbox{gray!50}{{0.9147}} & \colorbox{gray!50}{{5.3}} & \colorbox{gray!50}{0.0019} & \colorbox{gray!50}{0.9094} & \colorbox{gray!50}{5.1} & \colorbox{gray!50}{0.0074} & \colorbox{gray!50}{0.7992} & {5.3} & \colorbox{gray!50}{0.0076} & \colorbox{gray!50}{0.8224} & {6.1} & \colorbox{gray!50}{0.0120} & \colorbox{gray!50}{0.7443} & \colorbox{gray!50}{30.3} & \colorbox{gray!50}{0.0125} & \colorbox{gray!50}{0.7479} & \colorbox{gray!50}{32.1} \\
\tool-0.3   & \colorbox{gray!50}{0.0083} & \colorbox{gray!50}{0.8638} & \colorbox{gray!50}{\textbf{7.5}} & \colorbox{gray!50}{0.0066} & \colorbox{gray!50}{0.8768} & \colorbox{gray!50}{\textbf{8.6}} & \colorbox{gray!50}{0.0136} & \colorbox{gray!50}{0.6447} & \colorbox{gray!50}{\textbf{8.8}} & \colorbox{gray!50}{0.0148} & \colorbox{gray!50}{0.6601} & \colorbox{gray!50}{\textbf{8.9}} & \colorbox{gray!50}{0.0209} & \colorbox{gray!50}{0.5751} & \colorbox{gray!50}{\textbf{46.7}} & \colorbox{gray!50}{0.0208} & \colorbox{gray!50}{0.5831} & \colorbox{gray!50}{\textbf{41.3}} \\
Diffchaser & {0.0237}   & {0.6254}  & {2.6} & {0.0184}  & {0.6537} & {2.4} & {0.0391}  & {0.5328}  & {4.5} & {0.0581}  & {0.5219}  & {4.2} & {0.0864}  & {0.3283} & {6.1} & {0.0829} & {0.3211} & {2.9} \\
ADAPT & {0.0168} & {0.7212} & {3.7} & {0.0187} & {0.7502} & {3.4} & {0.0210}  & {0.5625}  & {8.1} & {0.0927} & {0.5542} & {8.2} & {0.0433} & {0.3827} & {25.9} & {0.0438} & {0.3949} & {23.1} \\
DLFuzz    & {0.0441}  & {0.5273}  & {1.2} & {0.0383}  & {0.6429}  & {1.2} & {0.0765} & {0.4933}  & {6.9} & {0.0513}  & {0.4521}  & {7.9} & {0.1037}   & {0.3283}   & {5.3} & {0.1047}  & {0.3028}     & {7.5} \\
DeepXplore & {0.0195} & {0.7091}  & {2.6} & {0.0181}  & {0.7248}  & {2.4} & {0.0223} & {0.5692}  & {7.4} & {0.0891} & {0.5182}  & {8.1} & {0.0538}   & {0.3972}   & {4.6} & {0.0529}  & {0.4101}   & {9.4} \\
\bottomrule
\end{tabular}
\caption*{\tiny \textit{Note:} Gray background highlights metrics where our method outperforms baselines; bold font indicates the best metric.}
\end{table}

\section{Results and Analysis}

\subsection{RQ1: How realistic are \tool\ generated samples?}

\subsubsection{Internal and External Comparison}
We conducted an internal comparison of \tool\ with different perturbation factor $\lambda$ and performed external comparisons with other baseline methods across six models on three datasets to evaluate the realism of the generated samples. We used two metrics to assess sample realism: MSE and SSIM, with detailed explanations of these metrics provided in Sec~\ref{metrics}. Lower MSE and higher SSIM values indicate that the generated adversarial samples have fewer differences from the original samples in terms of pixel values and structural similarity.

Table~\ref{sim} presents the results of comparing \tool\ with other baselines across three datasets using six models. To compare \tool\ with the baseline methods, we used gray background boxes to highlight all metrics where \tool\ outperforms the baselines and bold font to indicate the best metrics, making it easier to observe \tool's advantages. We show the results of \tool\ under different perturbation factors $\lambda$ (0.1, 0.2, 0.3), denoted as \tool-0.1, \tool-0.2, and \tool-0.3. Here, we chose the range of $\lambda$ primarily to maintain similarity between the generated samples and the original samples while introducing moderate variation. 

\tool\ demonstrates clear superiority in terms of MSE and SSIM. Specifically, for \tool-0.1, the highest similarity can be achieved, with the lowest MSE and the highest SSIM across all comparisons when $\lambda=0.1$. As the perturbation factor $\lambda$ increases from 0.1 to 0.3, a decrease in similarity is observed. However, even at $\lambda=0.3$, \tool\ still significantly outperforms all baselines in terms of similarity. As the perturbation factor increases, MSE rises and SSIM decreases, which is expected because a higher perturbation factor (e.g., $\lambda$ of 0.3) moves the sample further along the interpolation path toward the other data point. This results in greater divergence from the original sample.

Specifically, for MNIST, \tool-0.1 achieved an MSE of 0.0011 on LeNet4 and 0.0012 on LeNet5, with SSIM values of 0.9869 and 0.9902, respectively. This indicates that \tool\ generates highly realistic adversarial samples that closely similar the original ones. For CIFAR10, \tool\ continues to demonstrate excellent performance in terms of realism. Internally, the trends observed on CIFAR10 and ImageNet are similar to those on MNIST. In external comparisons, \tool-0.3 also outperforms all baselines in both MSE and SSIM.

\subsubsection{Manual Evaluation}

\begin{figure*}[h]
    \centering
    % 第一行: MNIST
    \begin{subfigure}[b]{0.45\textwidth}
        \centering
        \includegraphics[width=\textwidth]{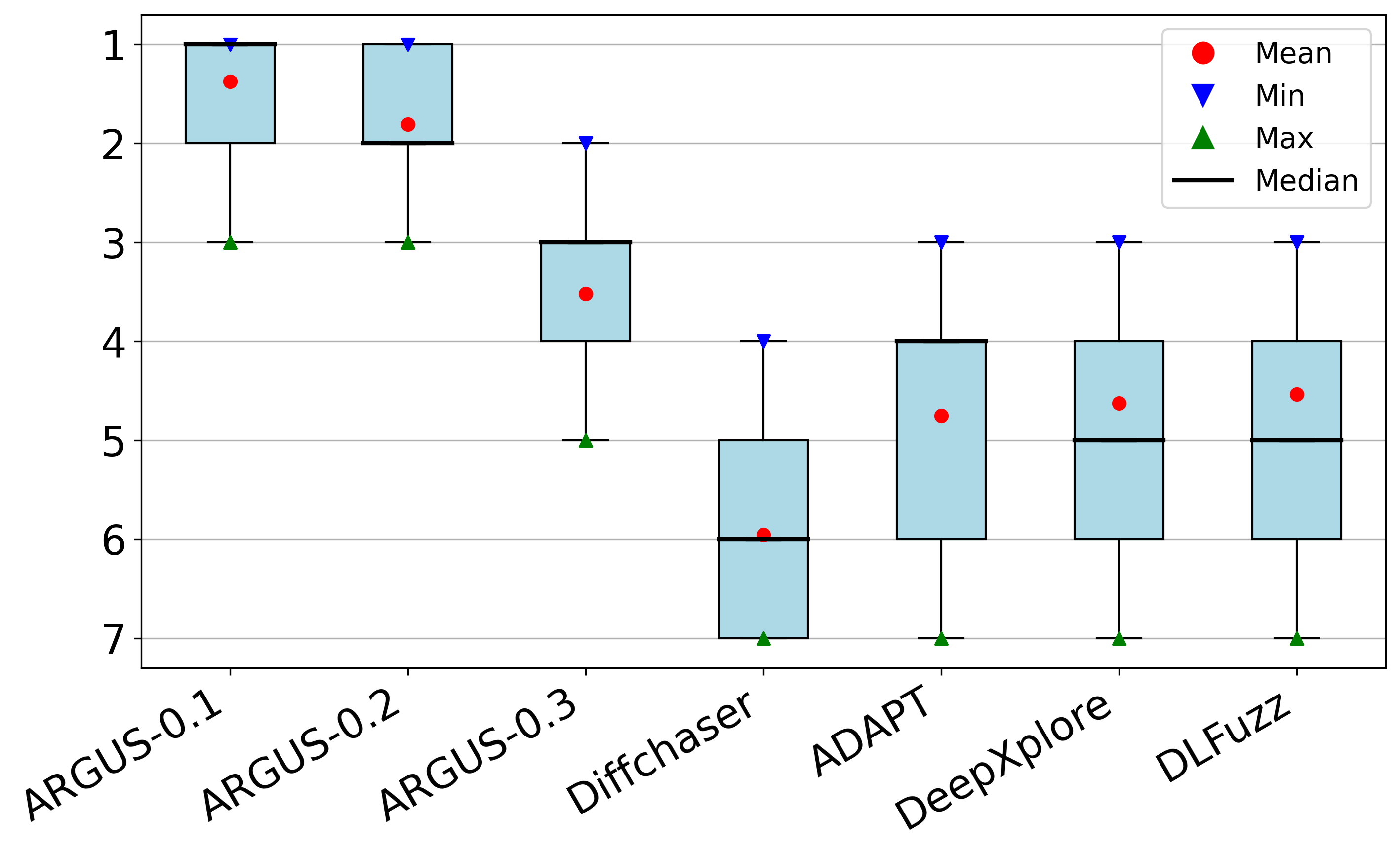}
        \caption{MNIST Rank}
    \end{subfigure}
    \begin{subfigure}[b]{0.45\textwidth}
        \centering
        \includegraphics[width=\textwidth]{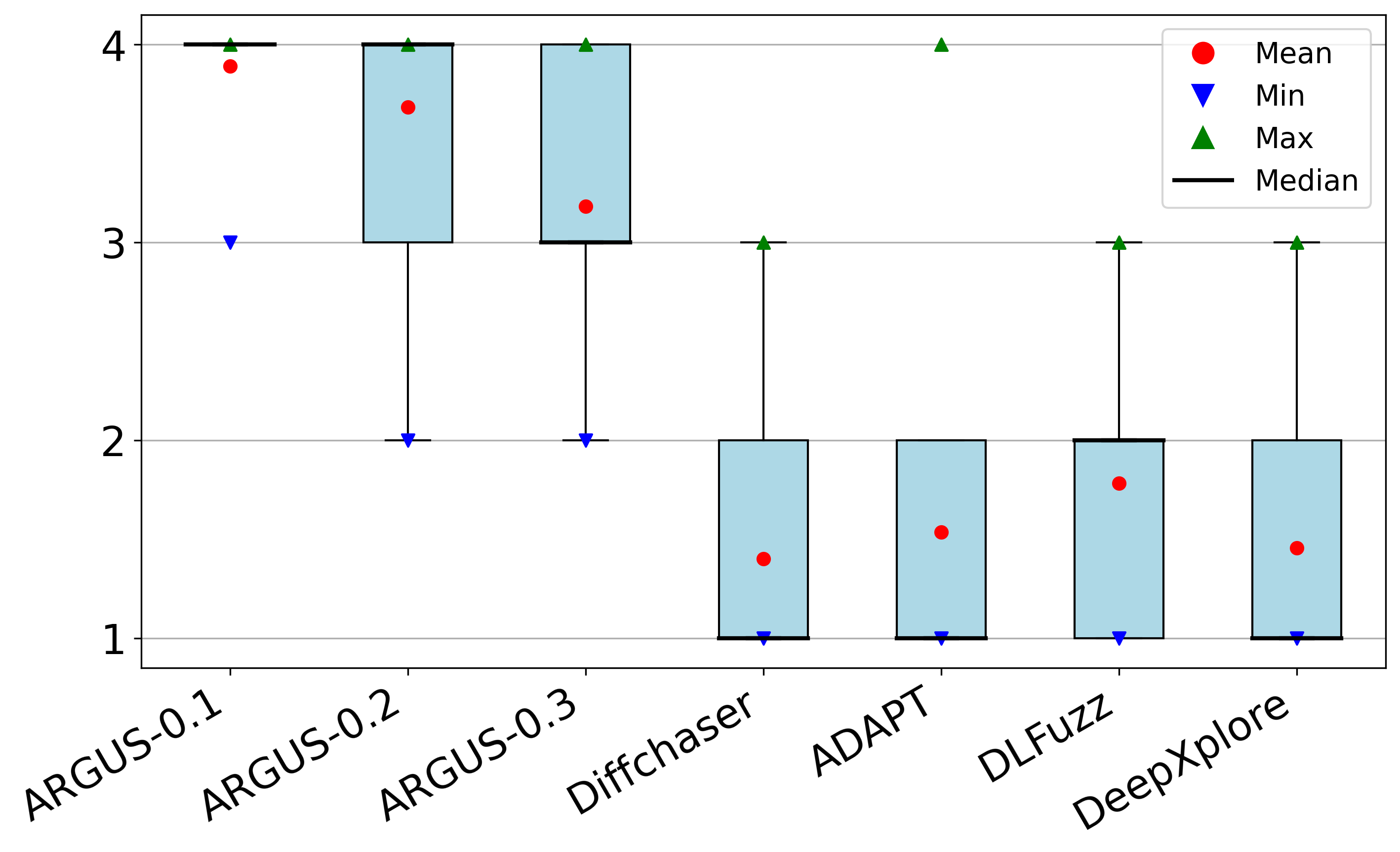}
        \caption{MNIST Score}
    \end{subfigure}
    \\
    % 第二行: CIFAR-10
    \begin{subfigure}[b]{0.45\textwidth}
        \centering
        \includegraphics[width=\textwidth]{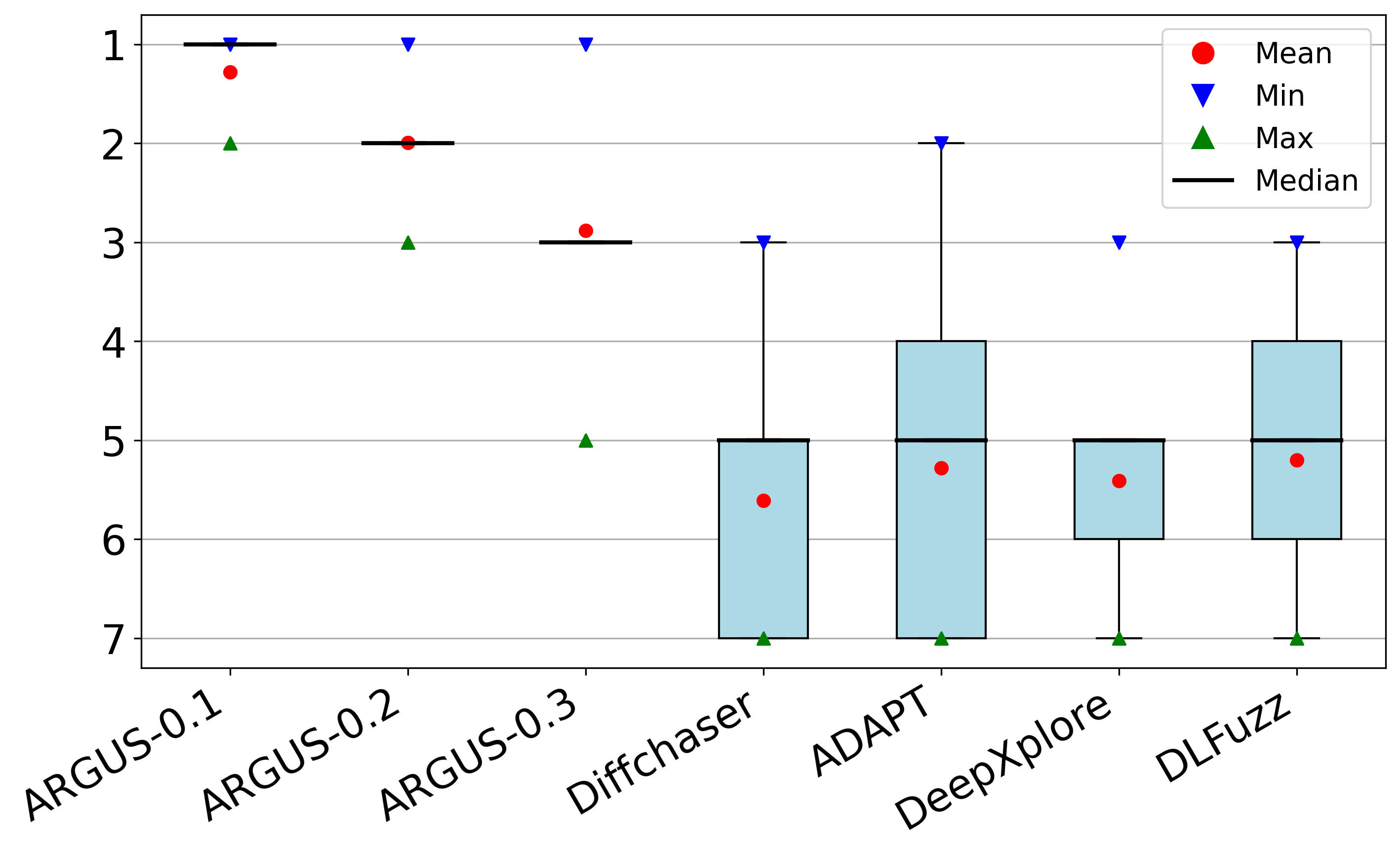}
        \caption{CIFAR10 Rank}
    \end{subfigure}
    \begin{subfigure}[b]{0.45\textwidth}
        \centering
        \includegraphics[width=\textwidth]{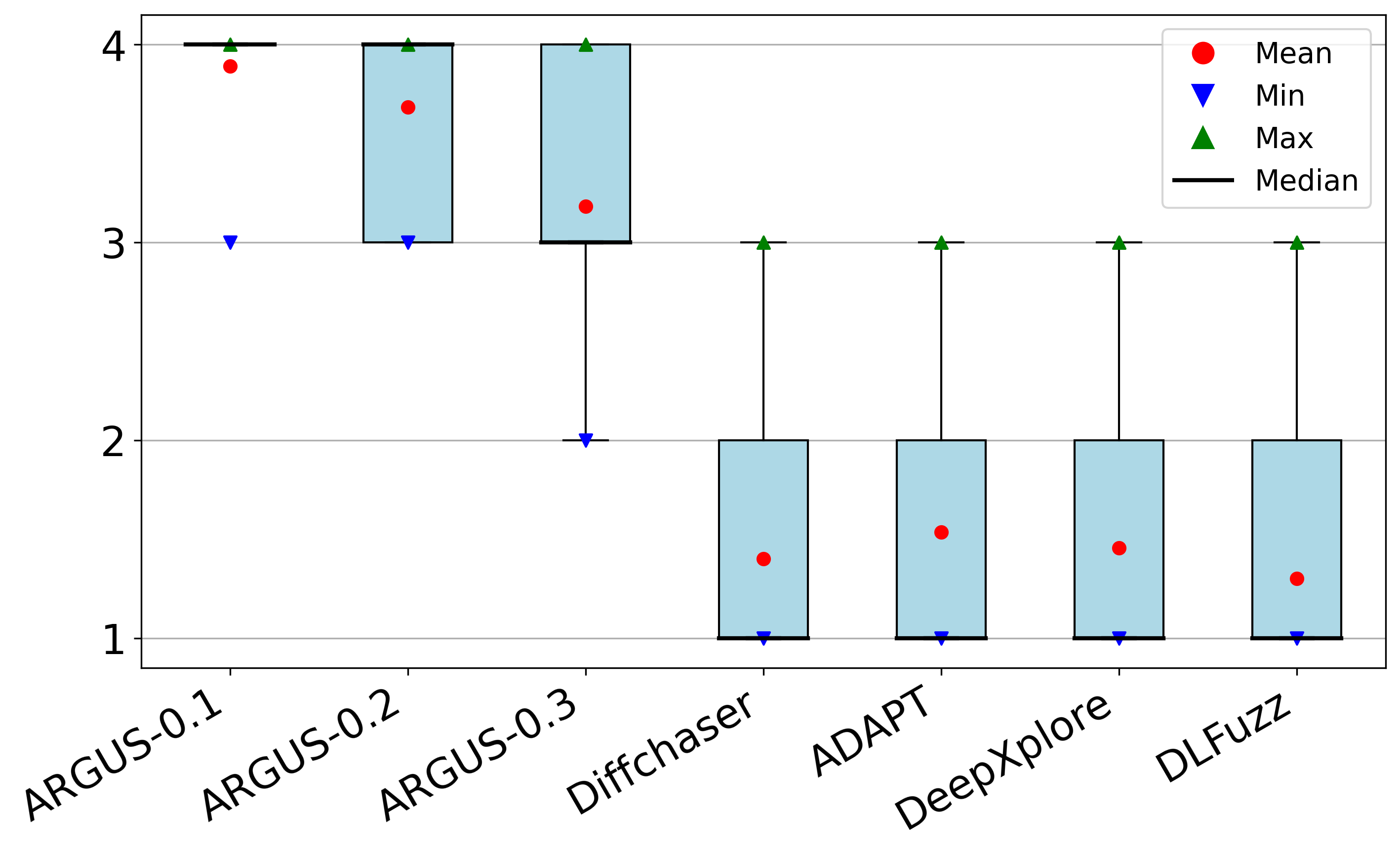}
        \caption{CIFAR10 Score}
    \end{subfigure}
    
    \caption{Results of Manually Evaluating Sample Realism.}
    \label{fig:comparison}
\end{figure*}

In addition to evaluating the sample realism with the two popular metrics MSE and SSIM, 
we also conducted a manual evaluation comparing adversarial images generated by \tool\ under different perturbation factors $\lambda$ (0.1, 0.2, 0.3) and four baseline methods on the CIFAR-10 and MNIST datasets. 
{ImageNet was excluded due to the challenges of this dataset, including its large number of categories and fine-grained class distinctions (e.g., dog breeds, flower types, etc.), which would make manual evaluation both cognitively demanding and time-consuming. Since participants are expected to  complete the current evaluation in approximately 20 minutes, including ImageNet would significantly increase their workload, potentially compromising the reliability of the results.}

We randomly selected one image per class, resulting in 10 original images per dataset, and for each of these images we randomly selected one corresponding adversarial sample generated by each of the seven methods under comparison. Therefore, in total there are 20 original samples and 140 adversarial samples. Ten people including 2 undergraduate, 3 master's, and 5 PhD students were recruited to carry out this evaluation. Each participant independently assesses the realism of adversarial samples in comparison to the original image from two perspectives: (1) Ranking, where adversarial samples are ordered from most to least similar (1 to 7), and (2) Scoring, where each sample is rated on a Four-Box Likert Scale (Excellent, Good, Fair, Poor), with "Excellent" corresponding to 4 points and "Poor" corresponding to 1 point. In this way, higher ranks and greater scores indicate better realism of the corresponding adversarial samples.

Fig~\ref{fig:comparison} shows the results of the manual evaluation. 
For MNIST, Argus-0.1's generated images are consistently considered as the most realistic, with the most compact Rank distribution centered near 1 as well as the greatest mean Score with a narrow interquartile range (IQR), which suggests strong agreement among evaluators. Argus-0.2 follows closely, maintaining a high Rank and high Score, demonstrating the relatively high realism of its generated adversarial samples. Argus-0.3 exhibits a wider Rank distribution and slightly lower Scores but still achieves an average Score above 3 (Good), indicating its effectiveness in generating high-realism adversarial samples. For CIFAR-10, the advantage of Argus is even more pronounced. Argus-0.1 is considered the best again according to the evaluation results, with the tightest distribution and the best mean value for both Rank and Score, reaffirming its superiority across evaluations. Argus-0.2 is the second, maintaining consistently high and stable Ranks and Scores, while Argus-0.3 significantly outperforms all baselines, with a mean Score above 3, confirming its ability to generate perceptually realistic adversarial samples.

In contrast, the adversarial images generated by baseline methods are considered to be less realistic according to the evaluation. Their Rank box plots exhibit worse mean and median values (closer to 5-7), while their Score distributions remain lower (mean values are mostly below 2) with wider IQRs, reflecting their inferior sample realism.

These results highlight the advantage of \tool\ over baseline methods in generating realistic adversarial samples, with \tool-0.1 consistently achieving the best performance, followed closely by \tool-0.2 and \tool-0.3.

\subsection{RQ2: How diverse are \tool\ generated samples?}

We evaluated the label diversity of samples generated by \tool\ through internal comparisons using different perturbation factor $\lambda$ and external comparisons with other baseline methods across six models and three datasets. We used the metric LD mentioned in Section~\ref{metrics}, where a higher LD indicates greater class diversity covered by the generated samples.

{Table~\ref{sim} presents the results of comparing \tool\ with other baselines across three datasets and six models. \tool\ demonstrates a significant advantage in terms of label diversity. With \tool-0.1, LD is relatively low, but it increases as $\lambda$ grows. %For \tool-0.4, nearly all classes are triggered across most datasets, which is an excellent result.
Notably, with \tool-0.3, LD reaches 7.5 for LeNet4 and 8.6 for LeNet5, which is more than double the values achieved by other baselines.
Among baselines, ADAPT performs best on Cifar10 and ImageNet, yet \tool\ still outperforms all baseline methods at $\lambda=0.3$ across all datasets and models. 

Considering the SSIM metric from RQ1, we observe some interesting trends:
when we increase $\lambda$ from 0.1 to 0.3, SSIM gradually decreases (0.99 $\rightarrow$ 0.87) while LD increases (3.6 $\rightarrow$ 8.6) on LeNet5, indicating that even with a large $\lambda$, adversarial samples remain highly similar to the originals. On the contrary, CIFAR10 is more sensitive to perturbations, with SSIM declining faster (0.91 $\rightarrow$ 0.66) while LD rising significantly (4.1 $\rightarrow$ 8.9), suggesting that even a small $\lambda$ value can induce diverse misclassifications. ImageNet exhibits the most drastic changes, with SSIM dropping sharply (0.89 $\rightarrow$ 0.58) while LD rising rapidly (18.2 $\rightarrow$ 41.3), reflecting the greater impact of perturbations in the latent space on high-complexity datasets. 
These findings suggest that MNIST's simpler structure makes it more resilient to this kind of perturbation, maintaining high SSIM while exhibiting a stable LD increase. In contrast, CIFAR10, with its more complex textures and color distributions, is more susceptible to perturbations, leading to a faster SSIM decline. The most pronounced shifts occur in ImageNet, indicating that even under moderate perturbations in the latent space, adversarial samples in high-dimensional datasets can significantly diverge from their originals.

From a trade-off perspective, an excessively small $\lambda$ retains high realism but limits the diversity of adversarial samples, reducing their effectiveness in testing, while a moderate increase of $\lambda$ can enhance sample diversity, contributing to more comprehensive testing. 
%A balance is needed: excessively high SSIM restricts adversarial variation, reducing effectiveness, while a moderate LD increase enhances diversity, leading to better evaluation. 
According to our experiments, the range $\lambda=0.2\sim0.3$ seems to achieve a good balance between realism and diversity of the generated adversarial samples.
%, making it a suitable reference for adversarial sample generation. 
That said, finding a suitable $\lambda$ should be dataset-specific, as the value for $\lambda$ depends on the dataset's complexity and sensitivity to perturbations.
}

\begin{table}[t]
\centering
\caption{Comparison of Model Test Accuracy Before and After Re-training}
\label{ICA}
\small
\begin{tabular}{@{}lccc@{}}
\toprule
& \multicolumn{1}{c}{MNIST} & \multicolumn{1}{c}{CIFAR10} & \multicolumn{1}{c}{ImageNet}\\
Samples    & LeNet5   & ResNet18  & ResNet50 \\
\midrule
Random      & 98.12--98.21 & 85.24--85.72 & \makecell{75.52–75.79 (Top-1) \\ 94.28–94.72 (Top-5)} \\
\hline
\tool-0.1   & 98.12--98.43 & 85.24--86.29 & \makecell{75.52–75.93 (Top-1) \\ 94.28–94.39 (Top-5)} \\
\hline
\tool-0.2   & 98.12--\textbf{99.28} & 85.24--87.46 & \makecell{75.52–76.21 (Top-1) \\ 94.28–95.29 (Top-5)} \\
\hline
\tool-0.3   & 98.12--99.18 & 85.24--\textbf{88.02} & \makecell{75.52–\textbf{76.47} (Top-1) \\ 94.28–\textbf{95.73} (Top-5)} \\
\bottomrule
\end{tabular}
\end{table}

\begin{table*}[t!]
\centering
\small
\caption{Comparison results of ER and SR}
\label{table:big_ER_SR}
\setlength{\tabcolsep}{2pt} 
\begin{tabular}{@{}lcc|cc|cc|cc|cc|cc@{}}
\toprule
& \multicolumn{4}{c}{MNIST} & \multicolumn{4}{c}{CIFAR10} & \multicolumn{4}{c}{ImageNet}\\
\cmidrule(lr){2-5} \cmidrule(lr){6-9} \cmidrule(lr){10-13}
& \multicolumn{2}{c}{LeNet4} & \multicolumn{2}{c}{LeNet5} & \multicolumn{2}{c}{VGG16} & \multicolumn{2}{c}{ResNet18} & \multicolumn{2}{c}{VGG19} & \multicolumn{2}{c}{ResNet50}\\
\cmidrule(lr){2-3} \cmidrule(lr){4-5} \cmidrule(lr){6-7} \cmidrule(lr){8-9} \cmidrule(lr){10-11} \cmidrule(lr){12-13}
Method    & ER    & SR    & ER    & SR    & ER    & SR    & ER    & SR     & ER    & SR    & ER    & SR   \\ 
\midrule
\tool-0.1   & \colorbox{gray!50}{23.48\%} & {100\%} & \colorbox{gray!50}{20.94\%} & {100\%} & {30.28\%} & {100\%} & {29.40\%} & {100\%} & \colorbox{gray!50}{38.13\%} & {100\%} & \colorbox{gray!50}{36.55\%} & \colorbox{gray!50}{{100\%}} \\
\tool-0.2   & \colorbox{gray!50}{32.24\%} & {100\%} & \colorbox{gray!50}{31.34\%} & {100\%} & {40.12\%} & {100\%} & \colorbox{gray!50}{39.96\%} & {100\%} & \colorbox{gray!50}{46.24\%} & {100\%} & \colorbox{gray!50}{49.23\%} & \colorbox{gray!50}{{100\%}} \\
\tool-0.3   & \colorbox{gray!50}{\textbf{45.56\%}} & {100\%} & \colorbox{gray!50}{\textbf{43.75\%}} & {100\%} & \colorbox{gray!50}{\textbf{55.13\%}} & {100\%} & \colorbox{gray!50}{\textbf{48.02\%}} & {100\%} & \colorbox{gray!50}{\textbf{61.37\%}} & {100\%} & \colorbox{gray!50}{\textbf{62.84\%}} & \colorbox{gray!50}{\textbf{100\%}} \\
Diffchaser & {2.51\%}   & {100\%} & {2.23\%}   & {100\%} & {10.93\%} & {100\%} & {2.23\%}  & {100\%}  & {5.20\%} & {43.5\%} & {1.30\%}  & {72.4\%} \\
ADAPT & {18.81\%}   & {100\%} & {19.62\%} & {100\%} & {43.77\%} & {100\%} & {38.93\%} & {100\%} & {14.92\%} & {100\%} & {18.03\%} & {94\%} \\
DLFuzz  & {11.24\%}   & {80\%}  & {8.35\%} & {90\%}  & {38.09\%} & {96\%}  & {39.13\%} & {100\%} & {8.93\%} & {92.4\%}  & {12.29\%} & {54.8\%} \\
DeepXplore & {12.55\%}   & {70\%}  & {14.23\%} & {85\%}  & {35.94\%} & {100\%} & {29.34\%} & {100\%} & {10.30\%} & {40.2\%} & {9.25\%} & {62.4\%} \\
\bottomrule
\end{tabular}
\end{table*}

\vspace{-3mm}
\subsection{RQ3: How effective are \tool\ generated samples in improving model performance?}

Here, we followed~\cite{pei2017deepxplore, wang2022bet}, and we tested whether the generated adversarial samples could improve model accuracy. We selected three different models from three different datasets, respectively, to conduct this experiment. We selected 1000 adversarial samples generated by \tool-0.1, \tool-0.2, and \tool-0.3 from the training sets of MNIST, CIFAR10, and ImageNet, as shown in table~\ref{ICA}. For MNIST and CIFAR10, this corresponds to 100 adversarial samples per class, and for ImageNet, we selected 20 adversarial samples per class. These adversarial samples were then mixed into the original training set. Subsequently, we fine-tuned the target models for 10 epochs and observed the change in accuracy on the test set. 

In order to compare the advantages of the generated samples, we also designed a comparative experiment, as shown in table~\ref{ICA}, named "Random". By ensuring the same number of samples in the training and test sets, we randomly removed 1,000 images from the test set, making sure that an equal number of samples were removed from each class. Under the same settings, these removed samples were added to the training set to form the training set for the comparative experiment, and the models were retrained accordingly.

As shown in Table~\ref{ICA}, re-training with samples generated by \tool-0.1, \tool-0.2, and \tool-0.3 improve test accuracy on MNIST, CIFAR10, and ImageNet. Specifically, \tool-0.3 achieves almost 3\% improvement on CIFAR10 and 1\% improvement on MNIST. For ImageNet, since the number of adversarial samples added per class is limited, the accuracy improvement is less pronounced compared to CIFAR10. However, both Top-1 and Top-5 accuracy show improvements. The improvement from \tool-0.1 is less significant than that from \tool-0.3. In contrast, in the “Random” comparison experiment, we observed that adding 1000 samples from the test set to the training set caused fluctuations in re-trained prediction accuracy, without the noticeable improvement seen with samples generated by \tool.

\subsection{RQ4: How effective are perturbations in \tool\ in modifying model output?}

We conducted an internal comparison of \tool\ and performed external comparisons with other baseline methods across six models on three datasets to evaluate the effectiveness of \tool. We used two metrics to assess the generated samples: ER and SR, with detailed explanations of these metrics provided in Sec~\ref{metrics}. A higher ER indicates that a greater proportion of the generated samples are effective adversarial samples. A higher SR demonstrates that the perturbations generated by \tool\ are more effective.

Table~\ref{table:big_ER_SR} presents the results of various methods on three datasets and six models. As the table shows, \tool\ demonstrates clear advantages across all perturbation levels. For all three datasets, the ER values of \tool\ consistently increase as the perturbation strength rises (from 0.1 to 0.3), indicating that \tool\ becomes more effective at causing misclassifications as the perturbation factor increases. 
In contrast, the ER values of other methods remain generally low. For example, in MNIST, the ER for DiffChaser, ADAPT, DLFuzz, and DeepXplore on LeNet4 all stay below 20\%, whereas the lowest ER for \tool\ reaches 23.48\%. On LeNet5, except for ADAPT, the ER for other methods remains similarly low, with ADAPT is 19.62\%, while \tool\ achieves a maximum ER of 43.75\%, which is approximately double that of ADAPT.

On the CIFAR10, ADAPT is the best baseline, reaching 43.77\% and 38.93\%. Other white-box testing methods show error rates between 35\% and 40\%, but \tool\ still exhibits a clear advantage on both VGG16 and ResNet18, achieving high ER of 55.13\% and 48.02\%, respectively. On the ImageNet dataset, the advantage of \tool\ becomes even more pronounced, with ER values approximately 3 times higher than those of the best baseline method, ADAPT.

From these results, it is evident that \tool\ achieves higher ER. Regardless of the variation of the data set or model, \tool-0.3 significantly outperforms other baseline methods, and the adversarial samples they generate are increasingly likely to induce classification errors as the perturbation factor increases. As for the SR metric, \tool\ maintains a 100\% success rate across all datasets, models, and perturbation factors, consistently finding adversarial inputs that mislead the model for all original samples.

\subsection{RQ5: Does each of the main components of \tool\ contribute to its effectiveness?}

Our generation method primarily relies on a combination of VQ-VAE and a dual discriminator structure. To verify the contribution of these components to the effectiveness of \tool, we conducted an ablation study. To control experimental cost, we picked one of the middle two values for $\lambda$, i.e. $\lambda = 0.2$ for this experiment.

%\begin{wraptable}{r}{0.5\textwidth}
\begin{table}[ht]
\centering
\caption{Performance Metrics for ResNet18 on CIFAR10}
\label{ablation}
\small
\begin{tabular}{@{}lccccc@{}}
\toprule
Method    & MSE    & SSIM   & LD  & ER    & SR    \\ 
\midrule
\tool-0.2       & \textbf{0.0076} & \textbf{0.8224} & \textbf{6.1} & \textbf{39.96\%} & \textbf{100\%} \\
\tool-0.2-VQ    & 0.0243 & 0.6594 & 6.0 & 32.28\% & 100\% \\
\tool-0.2-2D    & 0.0301 & 0.6971 & 6.1 & 36.94\% & 100\% \\
\bottomrule
\end{tabular}
\end{table}
%\end{wraptable}

The results are shown in Table~\ref{ablation}, we implement the ablation study on CIFAR10 using ResNet18. "\tool-0.2-VQ" represents \tool-0.2 without the vector quantizer in the latent space, using a standard VAE structure instead. "\tool-0.2-2D" represents \tool-0.2 to disable the dual discriminator, using only the VQ-VAE structure. As can be seen in the table, overall performance drops significantly when either component is removed, indicating that both parts are critical to the success of \tool. Additionally, it can be seen that disabling the vector quantizer and discriminator has little impact on LD, as changes in LD primarily stem from the interpolation perturbation method itself. However, since interpolation is core to achieving perturbation, it cannot be disabled. This indicates that the main factors influencing LD are the interpolation perturbation method and $\lambda$.

\section{Threats to Validity}

To mitigate threats to internal validity, we carefully reviewed the code and repeated the experiments three times to take the average to ensure the stability of the results. During model training, we used the default hyperparameter settings. For VGG and ResNet, we used PyTorch's pre-trained parameters, modifying only the output dimension of the final classifier. For the LeNet network, we strictly followed the structure and parameters specified in the original paper.
To mitigate threats to external validity, we reused all baseline methods using default hyperparameter settings. For the developed components, we manually checked the baseline results. Furthermore, in future work, we use three image classification datasets, and we plan to extend \tool\ to datasets in other domains.

\section{Related Work}

Recent testing methods fall into two categories: white-box and black-box testing, based on DNN transparency.  
Coverage-based methods are suited for white-box testing~\cite{pei2017deepxplore} as they require access to model details like weights and parameters, limiting their applicability in black-box testing, which lacks internal model access. Instead, black-box methods~\cite{odena2019tensorfuzz, xie2019deephunter} rely on DNN output feedback to infer errors and guide sample generation. While they cannot directly observe neuron activity, they detect anomalies in outputs to improve testing.  
Although neuron coverage-based white-box testing helps analyze DNN mechanisms, in cases where model details are inaccessible, black-box methods remain essential.

\subsection{White Box DNN Model Testing Methods}

DeepXplore~\cite{pei2017deepxplore} introduced neuron coverage to describe a model’s internal states, using gradient ascent to find error-inducing inputs and enhance coverage. DLFuzz~\cite{guo2018dlfuzz} applies fuzz-testing principles, updating the seed corpus and prioritizing samples based on coverage. DeepHunter~\cite{xie2019deephunter} generates perturbations via deformation mutations instead of gradient-based optimization. Adapt~\cite{lee2020effective} improves white-box testing by replacing fixed neuron selection with an online learning-based dynamic strategy. Robot~\cite{wang2021robot}, unlike methods focused on accuracy, strengthens model robustness against adversarial attacks.  

Overall, white-box testing methods prioritize high coverage but require deep model access, limiting their broader applicability.

\subsection{Black Box DNN Model Testing Methods}

TensorFuzz~\cite{odena2019tensorfuzz} pioneered black-box testing for DNNs using coverage-guided testing, where DNN outputs serve as coverage metrics. It relies on adding random noise to samples, which limits its ability to efficiently discover error-inducing inputs. DiffChaser~\cite{xie2019diffchaser} employs a genetic algorithm to detect inconsistencies between a DNN model and its compressed version.  
BET~\cite{wang2022bet} iteratively searches for error-inducing inputs by mutating original samples, evaluating modified samples in each iteration, and updating based on an objective function. ATOM~\cite{hu2023atom}, designed for multi-label classification, uses label combinations for test adequacy and leverages image search engines and NLP to find relevant test samples.  
These black-box methods rely heavily on extensive querying to identify error-inducing inputs, making them inefficient. Additionally, they operate in the input space, where only significant perturbations affect classification, often reducing the realism of the generated test inputs.  
{CIT4DNN~\cite{dola2024cit4dnn} introduces a novel method for generating diverse, error-revealing test samples by applying combinatorial interaction testing to the latent space of VAE. This approach leverages VAE’s reconstruction ability to ensure high similarity between generated and original samples, which inspires our work. 
However, CIT4DNN does not specifically target model weaknesses and instead focuses on generating rare inputs, which may not expose vulnerabilities.
}

\section{Conclusion}

In this paper, we introduced a black-box DNN model testing method, \tool, which aims to generate realistic, diverse, and fault-revealing test inputs. By performing interpolation in a continuous latent space, \tool\ perturbs the original samples using samples from other classes, then reconstructs them to the input space through quantization and decoding to obtain adversarial samples. 
Experimental results show that, compared to baseline methods, \tool\ excels in the similarity between generated samples and original samples and can cover more diverse labels. In terms of effective perturbation, \tool\ successfully perturbs all original samples and achieves a higher error rate in the model output. 
Furthermore, retraining with samples generated by \tool\ effectively improves the accuracy of the model.

\clearpage
\bibliographystyle{plain} 
\bibliography{main}

\end{document}